\documentclass[letterpaper, 10 pt, conference]{ieeeconf}
\usepackage{microtype}
\usepackage{graphicx}
\usepackage{booktabs}
\usepackage{amsmath, amssymb, amsfonts}
\usepackage{hyperref}
\usepackage{array}
\usepackage{cite}
\usepackage{multirow}
\usepackage{tabularx}

\IEEEoverridecommandlockouts

\title{\LARGE
Physical Contracts for Plug-Compatible Human--Humanoid Collaboration
}

\author{Hao Zhang$^{*,1,2}$, Yisen Li$^{*,1}$, Ruize Geng$^{2}$, Yves Tseng$^{1}$, Yaru Niu$^{2}$, Ding Zhao$^{\dagger,2}$ and H. Eric Tseng$^{\dagger,1}$
\thanks{$^{*}$Equal contribution.}
\thanks{$^{\dagger}$Correspondance to Ding Zhao (dingzhao@cmu.edu), H. Eric Tseng (hongtei.tseng@uta.edu) and Hao E. Zhang (haoz4@andrew.cmu.edu).}
\thanks{$^{1}$Hao Zhang, Yisen Li, Yves Tseng and H. Eric Tseng are with the University of Texas at Arlington.
        (email: haoz4@andrew.cmu.edu; yisen03@upenn.edu; ytseng@duck.com; hongtei.tseng@uta.edu)}%
\thanks{$^{2}$Hao Zhang, Ruize Geng, Yaru Niu and Ding Zhao are with Carnegie Mellon University
        (haoz4@andrew.cmu.edu; rgeng3@jh.edu; yarun@andrew.cmu.edu; dingzhao@cmu.edu)}%
}

\begin{document}

\maketitle

\thispagestyle{empty}
\pagestyle{empty}

\begin{abstract}
Full-stack human--robot collaboration (HRC) can become brittle when replacing a planner, partner model, coordination policy, or controller changes the physical meaning of cross-layer signals. We introduce C2C, an object-centric cognition-to-control architecture that preserves these meanings through physical contracts. Rather than standardizing individual modules, C2C standardizes the physical semantics exchanged between them: task intent is represented by a geometrically verified payload path, partner information by a source-independent physical state, learned coordination by bounded 11-D task-space commands, and robot-specific feasibility remains inside whole-body control (WBC). The reference system combines a vision-language model with deterministic geometric verification and multi-agent reinforcement learning (MARL) for partner-aware coordination. Across nine transport scenarios, adaptive MARL variants achieve 77.1--82.1\% mean success under unchanged interfaces, compared with 56.5\% for a scripted-partner reference. The same contracts further support neural and interpretable hard-tree actors, two- and three-carrier teams, and simulated-to-human partner substitution. Physical Unitree G1--human tests achieve 100\% success in spatially confined transport and 80\% in super-long-object handling, while a three-carrier system with two G1 humanoids and one human validates the same interface structure on hardware. Together, these results show that C2C converts a tightly coupled HRC stack into a plug-compatible system in which key components can change without redefining the physical collaboration task.
\end{abstract}

\section{Introduction}
\label{sec:introduction}

Human--humanoid collaborative transport is a demanding form of physical human--robot collaboration (HRC) because task reasoning, partner adaptation, and whole-body control must agree on the motion of one shared payload. Classical work established whole-body optimization for human--humanoid carrying \cite{agravante2019carrying,zhang2026halo}, while recent systems use haptic intent inference and hierarchical control \cite{bethala2025h2compact,zhang2025multi}, compliant reinforcement learning \cite{du2025cola}, or model-based planning with whole-body control \cite{kumbhar2025transport,zhang2026interaction}. These advances show that humanoids can collaborate with humans under substantial physical coupling. They also expose a systems problem: cognition, partner sensing, learned coordination, and robot execution are often connected through representations chosen for a particular stack.

\begin{figure}[t]
    \centering
    \includegraphics[width=0.88\columnwidth]{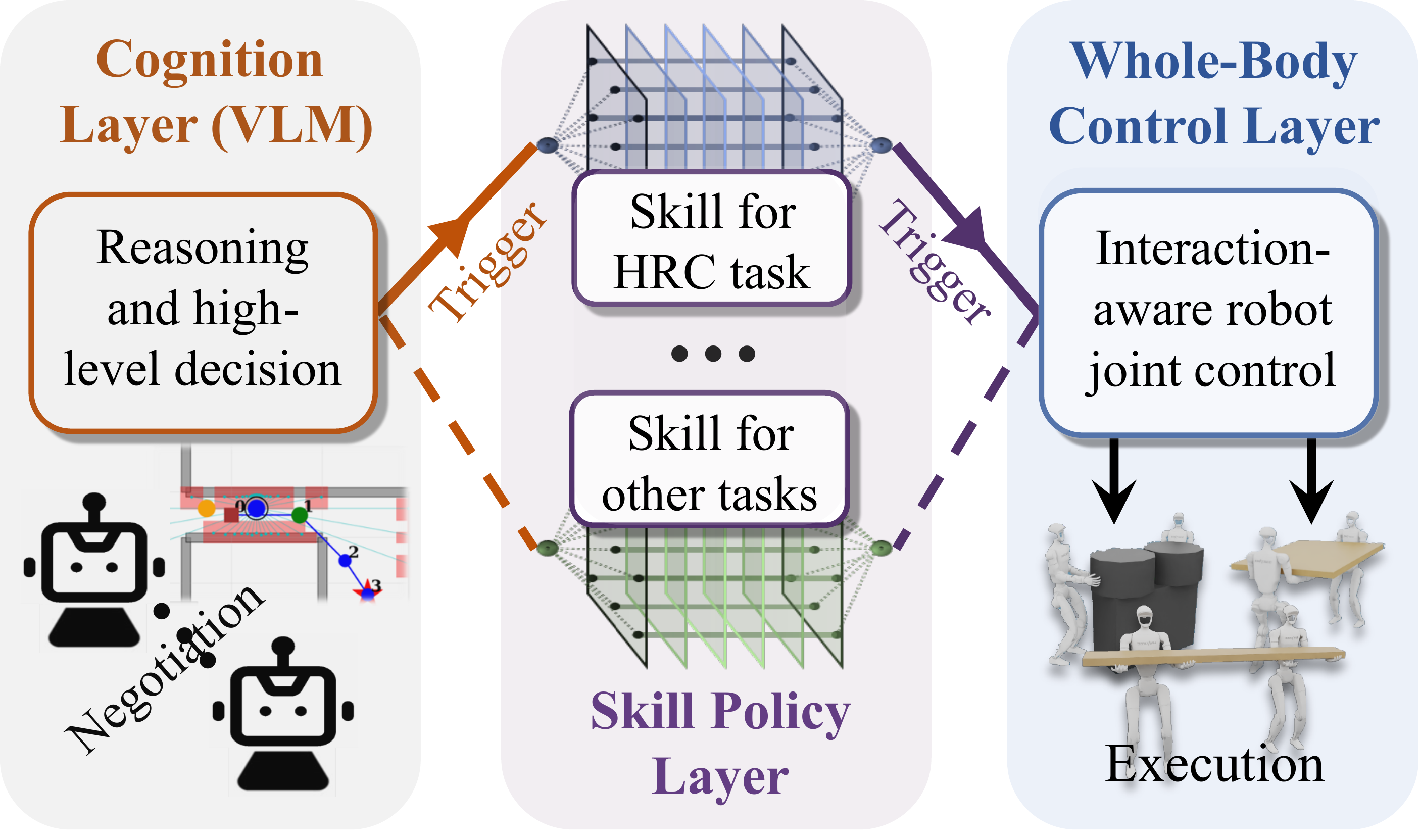}
    \caption{Cognition-to-control (C2C) resolves interface entanglement around the shared payload. Cognition outputs a verified object path, partner sensing is mapped to a source-independent physical state, coordination outputs bounded task-space adaptation, and robot-specific feasibility remains inside whole-body control. A module is plug compatible when it can change without changing these interface semantics.}
    \label{fig:concept_overview}
\end{figure}

The resulting failure mode is interface entanglement. A semantic planner may emit a skill identifier or planner-specific trajectory, a coordination policy may depend on simulator-specific partner states, and a low-level controller may require embodiment-specific references. Replacing any one component can therefore change the physical problem seen by its neighbors even when the software connection remains valid. This issue is becoming more important as semantic planners move closer to robot action \cite{ahn2022can,zhang2025bi} and humanoid controllers expose increasingly compact task-space interfaces \cite{dugar2025mhc,luo2026ceer,yang2026handoff}. Physical HRC adds a further constraint absent from single-robot modularity: the human and robot continuously perturb the same object, so task, partner, and execution semantics must remain mutually consistent. Fig.~\ref{fig:concept_overview} illustrates this cross-layer coupling.

We propose cognition-to-control (C2C), which treats the shared payload as the cross-layer physical invariant. Instead of standardizing internal models, C2C standardizes what each layer is allowed to expose: a verified payload path for task intent, a canonical physical state for the partner, a bounded task-space command for learned coordination, and robot-specific feasibility below the execution boundary. We call a substitution contract preserving when these quantities retain their physical meaning, units, and downstream responsibility after the module behind the boundary changes. The key design principle is therefore not software modularity alone, but preservation of physical semantics across cognition, interaction, and control. To the best of our knowledge, C2C is the first human--humanoid collaborative transport framework to treat cross-layer interface semantics as an explicit systems design objective and enforce them through physical contracts spanning cognition, partner sensing, learned coordination, and whole-body execution.

The contributions are summarized as follows: 1) We introduce object-centric physical contracts that define stable task, partner, coordination, and execution interfaces for full-stack HRC, localizing module changes without redefining the surrounding collaboration problem. 2) We realize these contracts in a human--humanoid transport system that connects verified vision-language guidance, partner-aware multi-agent reinforcement learning (MARL), and whole-body control (WBC) through explicit physical variables from payload path to joint torque. 3) We demonstrate contract-preserving substitution across coordination algorithms, policy representations, partner-state sources, and carrier counts, including executable hard-tree policies and multi-party human--humanoid collaboration.

\section{Related Work}
\label{sec:related_work}

\subsection{Human--robot collaborative transport}
Physical HRC has developed from compliant interaction and role-aware co-manipulation toward full-body collaborative transport \cite{ajoudani2018progress,ogenyi2021physical}. For humanoids, Agravante et al.\ formulated human--humanoid carrying as constrained whole-body optimization on HRP-4 \cite{agravante2019carrying}. Broader co-transport work has combined haptic and human-motion cues to adapt whole-body robot behavior \cite{sirintuna2022carrying}. More recent humanoid systems substantially expand physical capability: H$^2$-COMPACT decouples haptic intent interpretation from load-adaptive locomotion \cite{bethala2025h2compact}, COLA learns compliant leader--follower behavior in one policy \cite{du2025cola}, and Kumbhar et al.\ combine interaction-aware planning, compliance, and QP whole-body control on Digit \cite{kumbhar2025transport}. These works establish increasingly capable collaboration mechanisms. C2C targets a different systems question: how the physical task presented to neighboring modules can remain unchanged when the mechanism behind one layer is replaced. The distinction is between optimizing a collaboration stack and defining stable physical interfaces across the stack.

\subsection{From semantic planning to humanoid control}
Semantic robot planning has progressed from language-conditioned skill selection in SayCan \cite{ahn2022can} and closed-loop reasoning in Inner Monologue \cite{huang2022inner} to vision-language-action models such as RT-2 \cite{brohan2023rt2}. In parallel, humanoid control has moved toward reusable command abstractions. The Masked Humanoid Controller tracks partially specified target trajectories through a common whole-body interface \cite{dugar2025mhc}. CEER exposes compliant root and end-effector targets for hierarchical loco-manipulation \cite{luo2026ceer}, while HANDOFF uses a compact task-space command that can be driven directly by a VLM planner \cite{yang2026handoff}. These approaches motivate planner-friendly control interfaces, but collaborative transport introduces additional boundaries. The task representation must remain consistent with a shared payload, partner information must survive a change of sensing source, and learned coordination must not expose embodiment-specific commands upstream. C2C therefore extends the interface problem from planner--controller compatibility to cognition--partner--coordination--execution compatibility in a physically coupled team.

\subsection{Adaptive multi-agent coordination and partner substitution}
MARL provides a natural framework for coupled decision making, with centralized training and decentralized execution enabling agents to adapt to one another while retaining decentralized policies \cite{lowe2017multi,oroojlooy2023review}. HAPPO and HATRPO further address heterogeneous-agent policy updates through sequential and trust-region optimization \cite{zhong2022heterogeneous,kuba2021trust}. Cooperative learning with humans similarly emphasizes adaptation to partner behavior rather than fixed scripted responses \cite{hadfield2016cooperative}. Algorithmic adaptation alone, however, does not guarantee interface compatibility. A policy trained with a simulated teammate can depend on simulator-specific states, while a human partner is observed through measured physical signals. C2C makes the partner representation itself a contract, mapping simulated carriers and measured human motion into the same physical fields. The same principle is used to stress the coordination boundary with different optimizers, neural and hard-tree policies, and different carrier counts while preserving task and action semantics.

\begin{figure*}[t]
    \centering
    \includegraphics[width=0.999\textwidth]{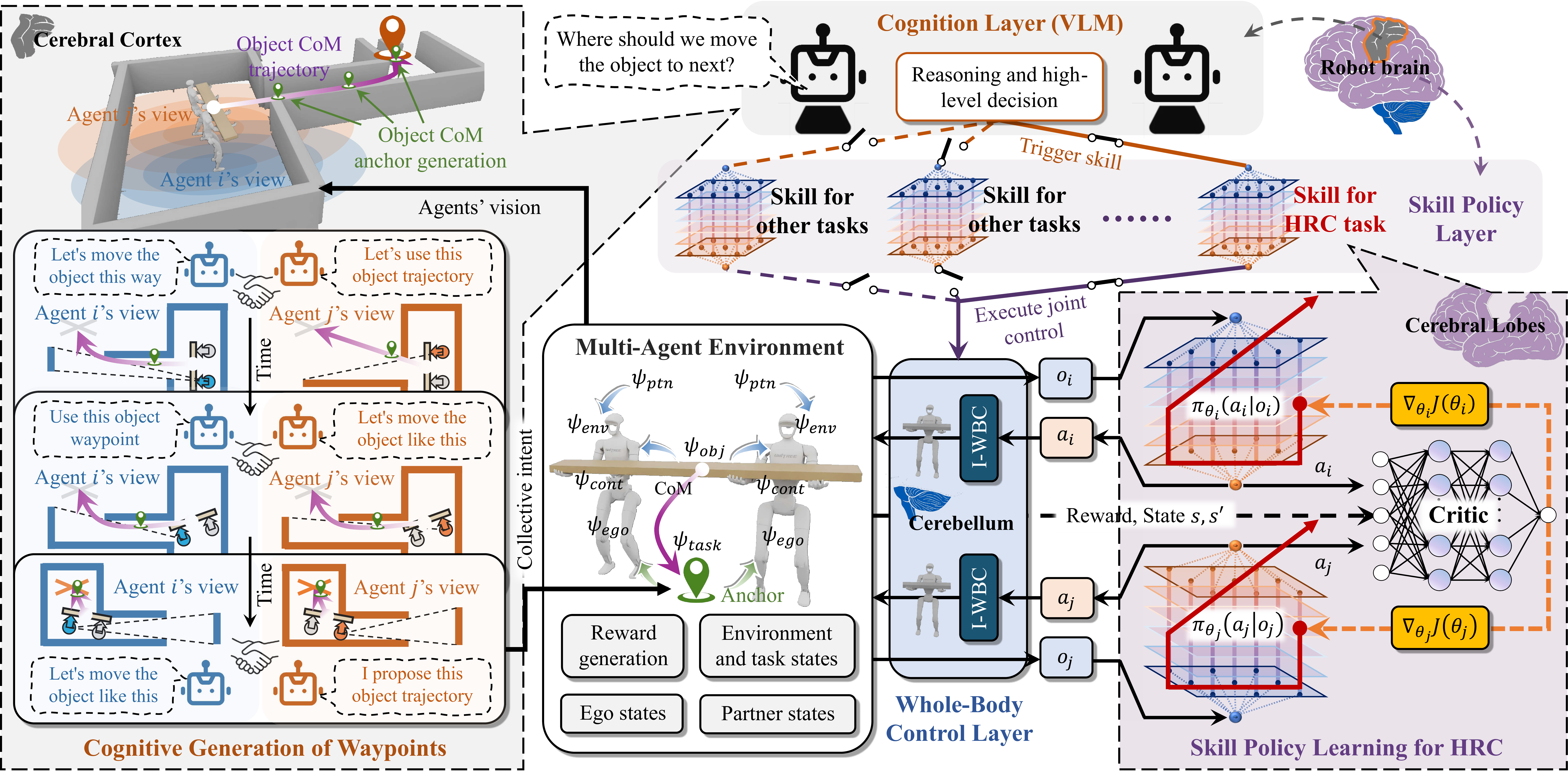}
    \caption{Reference realization of C2C. The two VLM views produce candidate paths for the same payload, not separate robot paths. A deterministic verifier fuses them into one object reference $P_t$. The coordination policy receives this reference together with self, partner, object, contact, and range states and outputs the same bounded 11-D task-space command $u_{i,t}$ for every tested policy. The execution layer converts that command to constrained joint torque. During hardware deployment, measured human motion fills the same partner-state fields used by the simulated teammate during training.}
    \label{fig:algorithm_framework}
\end{figure*}

\section{Contract-Preserving System Architecture}
\label{sec:methodology}

The architecture is defined by what crosses each boundary, not by the algorithm behind it. Fig.~\ref{fig:algorithm_framework} traces the data flow from cognition to joint torque. The four interfaces of the reference system are
\begin{equation}
\begin{aligned}
P_t &= f_{\rm cog}(I_t,g_t), & x^p_t &= f_{\rm p}(x^{\rm raw}_t),\\
u_{i,t} &= f_{\rm coord}(o_{i,t},P_t), & \tau_{i,t} &= f_{\rm wbc}(q_{i,t},\dot q_{i,t},u_{i,t}),
\end{aligned}
\label{eq:c2c_contract}
\end{equation}
where $P_t$ is the verified payload center-of-mass (CoM) path, $x^p_t$ is the canonical partner state, $u_{i,t}\in\mathbb{R}^{11}$ is the task-space command of carrier $i$, and $\tau_{i,t}$ is joint torque. Eq.~\eqref{eq:c2c_contract} defines each interface by physical meaning rather than internal representation. A replacement is contract preserving when meaning, units, and downstream responsibility remain unchanged. The experiments evaluate this requirement across the implemented substitutions.

The task contract ends cognition at an admissible object path rather than a language string or skill ID. The partner contract hides whether state comes from a simulated policy or a human sensor. The coordination contract limits learning to bounded task-space adaptation. The execution contract keeps dynamics, contacts, joint limits, and local clearance inside WBC. These boundaries stop implementation-specific representations before the next layer. A language string or skill ID would bind coordination to the planner vocabulary, a learned partner latent to a particular encoder, and joint targets to one robot morphology.

C2C instead terminates each module at a physical quantity needed by its neighbor. Planner internals end at $P_t$, sensing details at $x^p_t$, policy internals at $u_{i,t}$, and robot dynamics below WBC. This localization enables plug compatibility. The partner adapter is
\begin{equation}
x^p_t=f_{\rm p}(x^{\rm sim}_{j,t})
\quad\text{or}\quad
x^p_t=f_{\rm p}(\hat x^{\rm human}_{t}),
\label{eq:partner_contract}
\end{equation}
with identical field ordering, units, and normalization on both branches. Eq.~\eqref{eq:partner_contract} enables simulation-to-human substitution. At deployment the frozen robot policy receives measured human motion through the same partner interface.

\subsection{Verified VLM guidance}
\label{sec:vlm}
The left side of Fig.~\ref{fig:algorithm_framework} shows the cognition contract in the full stack. The reference cognition adapter uses Qwen2.5-VL-7B-Instruct \cite{bai2025qwen25vl} without task-specific fine-tuning. At each query, it receives one carrier-centered $64\times64$ top-down task raster for each carrier over an $8$~m$\times8$~m region. Each raster contains the payload footprint and CoM, goal, obstacles, collaborator locations, and numbered candidate anchors. These are structured task-state maps rather than raw camera images. C2C consumes the resulting task-state maps at the cognition boundary. Fig.~\ref{fig:sim_results}(a)--(c) shows the two view-conditioned proposals, geometric verification, and shared payload path.

Candidate anchors are generated before the VLM call. Obstacles are inflated by the payload footprint plus a 0.20~m margin. Free-space points are sampled on a 0.5~m grid, collision-pruned, and capped at 16 visible anchors. The fixed prompt is: ``Choose an ordered object-CoM path using only the numbered anchors shown in the image. Prefer progress toward the goal, payload clearance, and smooth heading changes. Return only the ordered anchor IDs. Do not create new coordinates.'' Decoding is deterministic and the parser accepts only displayed IDs.

The two calls use the same frozen model and prompt to produce view-conditioned proposals for the same payload, not independent robot plans. Both outputs are transformed to a common object frame before verification and selection. A proposal is admissible only when every segment remains inside the workspace and collision-free in every payload-inflated map. Let $C_t$ be this admissible candidate set. The selected object path is
\begin{equation}
P_t=\arg\min_{p\in C_t} L(p),
\label{eq:path_select}
\end{equation}
where $L(p)$ is path length and minimum clearance breaks ties. In Eq.~\eqref{eq:path_select}, the VLM proposes while deterministic geometry decides what crosses the task boundary. Malformed IDs, unavailable anchors, and invalid geometry trigger one re-query. If no valid proposal remains, deterministic shortest-path search produces the same $P_t$ on the fused occupancy map. If planning fails, nominal base motion stops.

Cognition runs at 0.5~Hz and is re-triggered when the active anchor is reached, the path becomes invalid, the goal changes, or payload deviation exceeds 0.35~m. Proposal, verification, and fallback all preserve the same $P_t$ contract.

The carriers do not receive independent paths. One payload pose generates compatible grasp targets. For carrier $i$,
\begin{equation}
p^d_{g_i}=p^d_o+R^d_o r_i,
\label{eq:grasp_target}
\end{equation}
where $p^d_o,R^d_o$ are the desired payload position and orientation and $r_i$ is the fixed grasp offset in the payload frame. Thus,
\begin{equation}
\|p^d_{g_i}-p^d_{g_j}\|_2=\|r_i-r_j\|_2,
\label{eq:rigid_distance}
\end{equation}
so Eq.~\eqref{eq:grasp_target} and Eq.~\eqref{eq:rigid_distance} derive every carrier target from one payload pose and fixed grasp offsets. The execution layer monitors measured grasp geometry and local clearance, stopping motion outside admissible bounds. Shared-payload geometry and WBC clearance constraints therefore preserve rigid-object consistency during execution.

\subsection{Partner-aware MARL behind a fixed task-space contract}
\label{sec:skill_policy}
The coordination policy is trained in Isaac Lab with two policy-controlled carriers under centralized training and decentralized execution (CTDE). During physical deployment, only the robot actor is executed. The human is a measured partner source and does not participate in gradient updates. MARL exposes the robot to an adaptive teammate during training, while the partner contract allows human substitution without changing the robot policy interface.

For each carrier, the current interaction block and final policy observation are
\begin{equation}
\begin{aligned}
h_{i,t}&=[x^{\rm task}_{i,t},x^{\rm self}_{i,t},x^{\rm partner}_{i,t},x^{\rm object}_{t},x^{\rm contact}_{t}]\in\mathbb{R}^{58},\\
o_{i,t}&=[h_{i,t},h_{i,t-1},h_{i,t-2},r_{i,t}]\in\mathbb{R}^{210},
\end{aligned}
\label{eq:observation}
\end{equation}
where the blocks have dimensions 10, 13, 13, 18, and 4, and $r_{i,t}\in\mathbb{R}^{36}$ is the current range vector. Self and partner fields contain planar pose and velocity, yaw, CoM height, torso pitch, and relative end-effector features. On hardware, motion capture reconstructs the partner fields. Object state contains payload pose, velocity, and four corners. Contact uses four binary end-effector indicators.

Eq.~\eqref{eq:observation} combines the current block with two previous snapshots for short history without recurrence. The 36-D environment channel is a radial proximity scan normalized as $1-\min(d,4.0)/4.0$. For active path anchor $j$, the 10-D task block is the next five relative payload waypoints,
\begin{equation}
x^{\rm task}_{i,t}=[w_j-p_{o,t},\ldots,w_{j+4}-p_{o,t}]\in\mathbb{R}^{10},
\label{eq:lookahead}
\end{equation}
with missing entries padded by the final anchor. Eq.~\eqref{eq:lookahead} localizes the payload contract by expressing exteroceptive quantities in the carrier frame.

The actor outputs $a_{i,t}\in[-1,\,1]^{11}$ and modulates only a nominal task-space command,
\begin{equation}
u_{i,t}=u^{\rm nom}_{i,t}+D a_{i,t},
\label{eq:residual_action}
\end{equation}
where diagonal $D$ sets fixed physical residual limits. Eq.~\eqref{eq:residual_action} keeps learned adaptation in task space and avoids unrestricted joint-level control. The command itself is
\begin{equation}
u_{i,t}=[v_x,v_y,\omega, H_{\rm CoM},\theta_{\rm torso},p^L_w,p^R_w]^T,
\label{eq:command_vector}
\end{equation}
with three planar base commands, CoM height, torso pitch, and two 3-D wrist targets. Eq.~\eqref{eq:command_vector} and Table~\ref{tab:io_spec} summarize the policy input and output. The planner sets object motion, MARL adapts to the partner, and WBC executes the command.

The shared task reward is
\begin{equation}
R_t=w_p\,\Delta s_t-w_l\sum_{c=1}^{4}|z_c-\bar z|-w_d I_{\rm drop},
\label{eq:reward}
\end{equation}
where $\Delta s_t$ is signed progress along the verified payload path, the second term penalizes tilt through the four corner heights, and $I_{\rm drop}$ marks grasp loss or payload drop. Progress is credited only inside the 0.35~m path corridor. The leveling term is disabled for pushing tasks where payload leveling is not required. Eq.~\eqref{eq:reward} is shared by all coordination methods.

A centralized critic is used only during training. For proximal policy optimization (PPO)-based variants, actor $i$ uses
\begin{equation}
L_i=\mathbb{E}\left[\min\left(r_{i,t}A_t,\,\operatorname{clip}(r_{i,t},1-c,1+c)A_t\right)\right],
\label{eq:ppo_objective}
\end{equation}
where $r_{i,t}$ is the new/old policy likelihood ratio, $A_t$ is the centralized GAE advantage, and $c$ is the clip bound. Eq.~\eqref{eq:ppo_objective} defines the PPO-style local update used by the corresponding variants. HAPPO retains its sequential update \cite{zhong2022heterogeneous}, HATRPO its trust-region update \cite{kuba2021trust}, and PPO+PCGrad projects conflicting gradients in the shared interaction encoder \cite{yu2020gradient}. These optimizer-specific updates remain inside coordination. Observation, reward, and the 11-D output contract stay fixed.

\begin{table}[t]
\centering
\caption{Default two-carrier policy input and output.}
\label{tab:io_spec}
\scriptsize
\setlength{\tabcolsep}{2.4pt}
\renewcommand{\arraystretch}{1.08}
\begin{tabular*}{\columnwidth}{@{\extracolsep{\fill}} lcl}
\toprule
\textrm{Component} & \textrm{Size} & \textrm{Description} \\
\midrule
\multicolumn{3}{l}{Observation space ($o \in \mathbb{R}^{210}$)} \\
Global guidance & 10 & 5 anchors (relative planar) \\
Self state      & 13 & Pose/velocity, CoM, torso, wrists \\
Partner state   & 13 & Measured/policy partner \\
Object geometry & 18 & 4 corners, CoM pose/velocity \\
Contact         & 4  & 4 binary contact indicators \\
Temporal buffer & 116 & Two previous 58-D blocks \\
Environment     & 36 & Current 36-ray range \\
\midrule
\multicolumn{3}{l}{Action space ($a \in \mathbb{R}^{11}$)} \\
Locomotion      & 3  & Planar velocity/yaw residual \\
Postural        & 2  & CoM/torso residual \\
Wrist control   & 6  & Wrist-position residual \\
\bottomrule
\end{tabular*}
\end{table}

\begin{figure*}[t]
    \centering
    \includegraphics[width=0.99\textwidth]{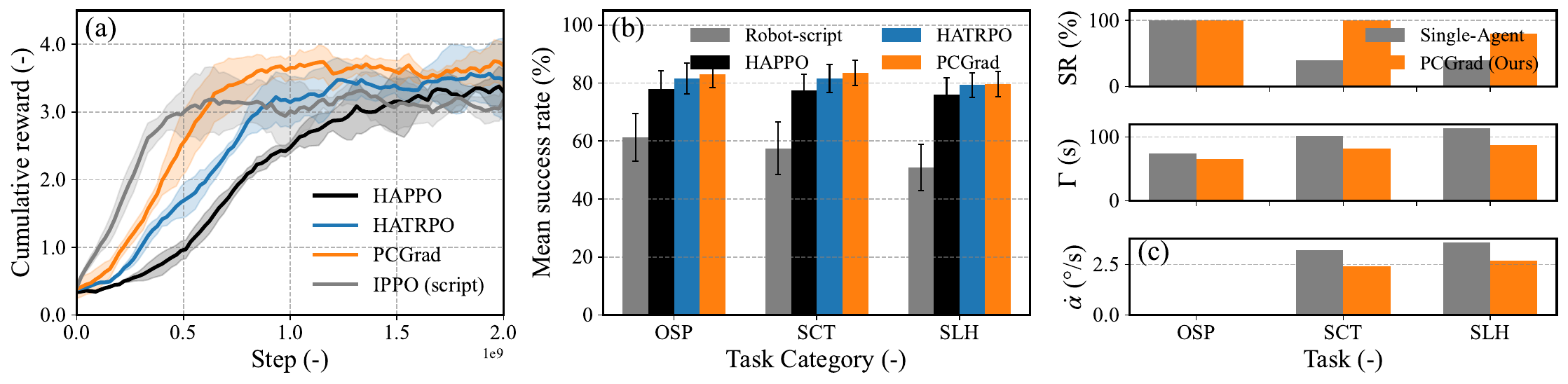}
    \caption{
        (a)~Episode return during training for IPPO with a scripted partner, HAPPO, HATRPO, and PPO+PCGrad over $2.0 \times 10^9$ steps, with shading denoting one standard deviation across five seeds.
        (b)~Mean success rate by task category, comparing the robot-script baseline with the same learning-based tactical variants. (c)~Physical deployment: success rate, completion time (s), and mean object tilt rate ($^\circ$/s) for robot-script IPPO and PPO+PCGrad.
    }
    \label{fig:return_and_tables}
\end{figure*}

\begin{table*}[t]
\caption{System-level success rate under the shared C2C interfaces across nine scenarios (mean $\pm$ standard deviation over five seeds, 100 episodes per seed). $\Delta$: relative difference of the mean adaptive-MARL success rate from the robot-script reference.}
\label{tab:giant_matrix}
\centering
\footnotesize
\renewcommand{\arraystretch}{1.15}
\begin{tabular*}{\textwidth}{@{\extracolsep{\fill}} ll ccccc}
\toprule
\textrm{Category} & \textrm{Scenario} & \textrm{Robot-script} & \textrm{HAPPO} & \textrm{HATRPO} & \textrm{PPO+PCGrad} & \textrm{Mean gain $\Delta$} \\
\midrule
\multirow{3}{*}{OSP}
& $S_{11}$: Alignment      & 65.4 $\pm$ 7.2 & 83.6 $\pm$ 5.3 & 87.9 $\pm$ 4.5 & 87.7 $\pm$ 3.8 & +32.1\% \\
& $S_{12}$: Turnaround     & 60.1 $\pm$ 8.5 & 77.6 $\pm$ 6.3 & 81.5 $\pm$ 5.5 & 83.6 $\pm$ 4.8 & +34.6\% \\
& $S_{13}$: Corner entry   & 58.2 $\pm$ 9.1 & 72.7 $\pm$ 7.0 & 75.3 $\pm$ 6.0 & 77.8 $\pm$ 5.3 & +29.3\% \\
\midrule
\multirow{3}{*}{SCT}
& $S_{21}$: Narrow gate    & 59.2 $\pm$ 9.0 & 80.1 $\pm$ 4.8 & 83.4 $\pm$ 4.3 & 88.6 $\pm$ 3.5 & +42.0\% \\
& $S_{22}$: S-shaped path  & 57.5 $\pm$ 8.8 & 76.3 $\pm$ 5.8 & 82.1 $\pm$ 5.0 & 80.6 $\pm$ 4.5 & +38.6\% \\
& $S_{23}$: U-shaped path  & 55.6 $\pm$ 9.5 & 75.7 $\pm$ 6.5 & 79.2 $\pm$ 5.3 & 81.3 $\pm$ 4.8 & +41.6\% \\
\midrule
\multirow{3}{*}{SLH}
& $S_{31}$: Facing mode    & 52.8 $\pm$ 8.1 & 79.3 $\pm$ 5.0 & 84.4 $\pm$ 1.6 & 83.2 $\pm$ 3.5 & +55.9\% \\
& $S_{32}$: Lateral shuffle & 50.4 $\pm$ 7.5 & 73.7 $\pm$ 6.8 & 75.9 $\pm$ 5.8 & 77.3 $\pm$ 5.0 & +50.1\% \\
& $S_{33}$: Pivoting       & 49.6 $\pm$ 8.3 & 74.9 $\pm$ 6.0 & 77.5 $\pm$ 5.3 & 78.6 $\pm$ 4.5 & +55.2\% \\
\midrule
\multicolumn{2}{l}{Mean success rate} & 56.5\% & 77.1\% & 80.8\% & 82.1\% & +41.5\% \\
\bottomrule
\end{tabular*}
\end{table*}

\subsection{Whole-body execution and conversion to joint torque}
The center branch in Fig.~\ref{fig:algorithm_framework} realizes the execution contract. The execution layer resolves the 11-D command at 100~Hz while the 2~Hz coordination command is held by zero-order hold between policy updates. For each base, CoM, torso, or wrist task, Eq.~\eqref{eq:task_feedback} generates the desired task acceleration,
\begin{equation}
\ddot x^d=\ddot x_{ff}+K_p(x^d-x)+K_d(\dot x^d-\dot x),
\label{eq:task_feedback}
\end{equation}
and all tasks are solved together by an inverse-dynamics quadratic program (QP),
\begin{equation}
\begin{aligned}
\min_{\ddot q,\tau,f_c}\,&\|J\ddot q+\dot J\dot q-\ddot x^d\|_W^2+w_\tau\|\tau\|_2^2\\
\text{s.t. }\,&M(q)\ddot q+h=S^T\tau+J_c^T f_c,\\
&J_c\ddot q+\dot J_c\dot q=0,\quad f_c\in K_f,\\
&\tau_{\min}\le\tau\le\tau_{\max},\quad q_{\min}\le q+\Delta q\le q_{\max},\\
&d(q)+J_d(q)\Delta q\ge d_{\rm safe}.
\end{aligned}
\label{eq:wbc_qp}
\end{equation}
In Eq.~\eqref{eq:wbc_qp}, $M,h$ are robot dynamics, $J,J_c$ are task/contact Jacobians, $S$ is the actuation selector, $f_c$ is contact force, $K_f$ is the friction-cone approximation, $W$ contains task weights, and $J_d$ is the clearance Jacobian. The one-step prediction is $\Delta q=\Delta t\,\dot q+\tfrac{1}{2}\Delta t^2\ddot q$.

The QP enforces rigid-body dynamics, contact consistency, friction, actuator and joint limits, and robot--obstacle/partner clearance. High weight on common-payload wrist references prevents the bounded MARL residual from breaking rigid grasp geometry.

The optimized torque is combined with joint feedback and saturation,
\begin{equation}
\tau_{\rm cmd}=\operatorname{sat}_{[\tau_{\min},\tau_{\max}]}\left(\tau^*+K_q(q^d-q)-K_v\dot q\right).
\label{eq:torque_command}
\end{equation}
If the QP is infeasible or measured clearance falls below 0.15~m, nominal base motion is zeroed and the last feasible posture is held. Eq.~\eqref{eq:task_feedback}--\eqref{eq:torque_command} therefore map the C2C command to joint torque while enforcing robot-specific feasibility.

\begin{table}[h]
\centering
\caption{Shared training settings.}
\label{tab:training_hyper}
\scriptsize
\setlength{\tabcolsep}{2.6pt}
\renewcommand{\arraystretch}{1.03}
\begin{tabular*}{\columnwidth}{@{\extracolsep{\fill}} l c l c}
\toprule
Parameter & Value & Parameter & Value \\
\midrule
Learning rate & $10^{-4}$ & Hidden layers & [256,256,128] \\
Epochs/update & 10 & Activation & ReLU \\
Minibatches & 16 & Initialization & Orthogonal \\
Entropy weight & 0.01 & Optimizer & Adam \\
Discount factor & 0.99 & Weight decay & $10^{-4}$ \\
GAE factor & 0.95 & Gradient clip & 10.0 \\
PPO/HAPPO clip & 0.2 & Learning-rate schedule & Cosine \\
Value-loss weight & 0.5 & Action steps & $2\times10^9$ \\
\bottomrule
\end{tabular*}
\vspace{-1mm}
\end{table}

\section{Validation of Contract Preservation}
\label{sec:experiments}

We evaluate four substitutions behind C2C boundaries: coordination algorithm, partner-state source, policy representation, and carrier count. Payload reference, observation semantics, 11-D task-space command, reward, and robot-side execution remain fixed. Table~\ref{tab:contract_evidence} maps each substitution to the preserved semantics and evidence.

The final row verifies end-to-end dependency. Removing either the guidance source or learned coordination breaks a required boundary on $S_{33}$, while the complete stack reaches 78.6\% SR in 81.2~s.

\subsection{Experimental setup}
\label{sec:setup}

\subsubsection{Task suite}
\begin{figure*}[t]
    \centering
    \includegraphics[width=0.95\textwidth]{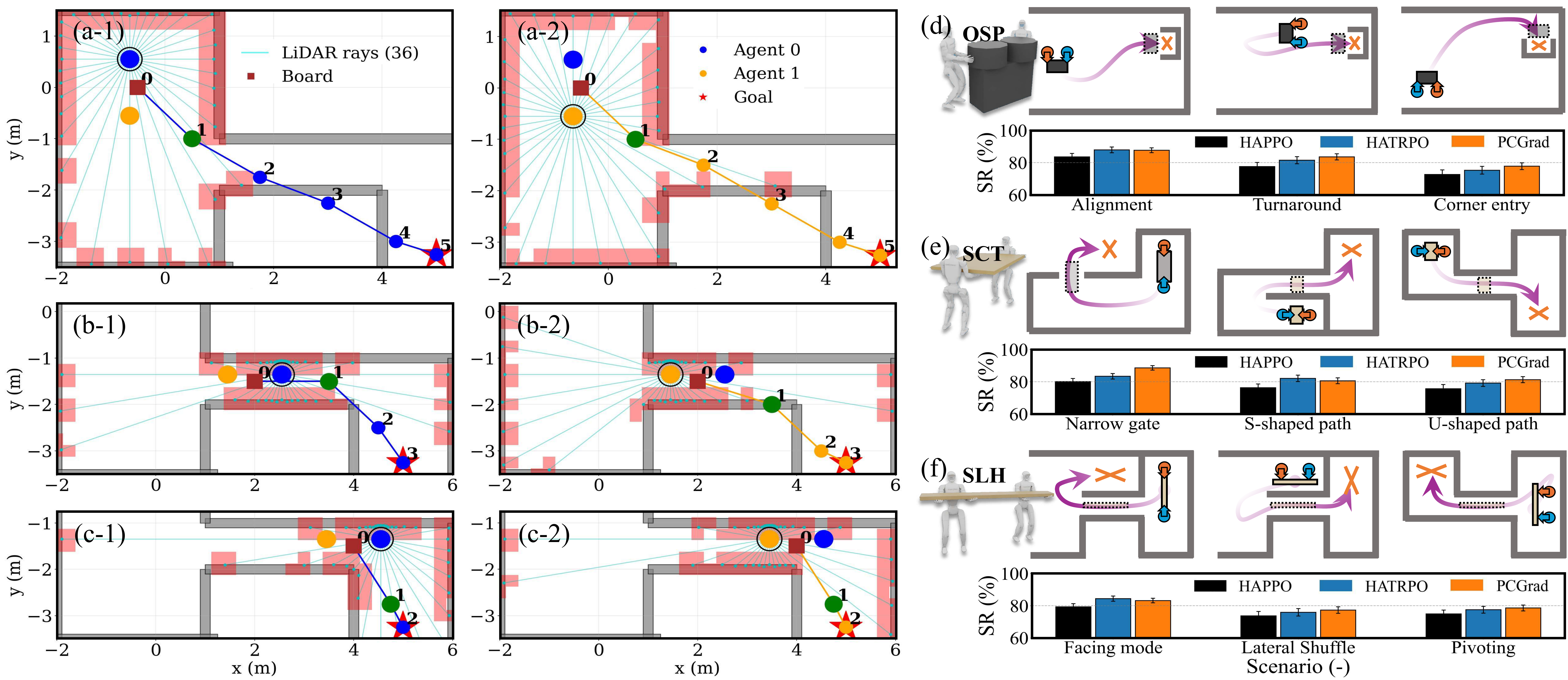}
    \caption{VLM guidance and system stress tasks. Panels (a)--(c) show two carrier-centered views. Blue and orange traces are payload-path proposals from the VLM adapter. Deterministic common-frame collision and clearance checks select the verified path, with the active anchor shown in green and the 36-D range observations in cyan. Panels (d)--(f) show the OSP, SCT, and SLH task families used under the same C2C contracts.}
    \label{fig:sim_results}
\end{figure*}

We use nine scenarios as a system stress suite. Orientation-sensitive pushing (OSP) contains $S_{11}$ alignment, $S_{12}$ turnaround, and $S_{13}$ corner entry. Spatially confined transport (SCT) contains $S_{21}$ narrow gate, $S_{22}$ S-shaped path, and $S_{23}$ U-shaped path. Super-long-object handling (SLH) contains $S_{31}$ facing mode, $S_{32}$ lateral shuffle, and $S_{33}$ pivoting. They respectively stress yaw alignment, synchronized confined motion, and long-payload stabilization while sharing the observation, residual-action, reward, and WBC contract. Fig.~\ref{fig:sim_results}(d)--(f) shows the three task families.

\begin{table*}[t]
\caption{Summary of contract-preserving substitutions under the C2C interfaces.}
\label{tab:contract_evidence}
\centering
\scriptsize
\setlength{\tabcolsep}{4.5pt}
\renewcommand{\arraystretch}{1.12}
\begin{tabularx}{\textwidth}{
    >{\raggedright\arraybackslash}p{0.12\textwidth}
    >{\raggedright\arraybackslash}p{0.25\textwidth}
    >{\raggedright\arraybackslash}p{0.31\textwidth}
    >{\raggedright\arraybackslash}X}
\toprule
Axis & Substitution & Invariant interface & Evidence \\
\midrule
Cognition
& VLM $\rightarrow$ deterministic fallback
& Verified payload path $P_t$
& Same downstream path contract \\

Coordination
& HAPPO / HATRPO / PPO+PCGrad
& Observation, 11-D action, reward, WBC
& SR: 77.1 / 80.8 / 82.1\% \\

Partner
& Simulated carrier $\rightarrow$ human
& Partner-state fields and robot action
& Physical SR: 100 / 100 / 80\% \\

Policy
& MLP $\rightarrow$ hard tree
& N=3 adapter, action, reward, WBC
& SR: 74.7 $\rightarrow$ 70.3\% \\

Team size
& Two $\rightarrow$ three carriers
& Guidance, action semantics, reward, WBC
& SR: 77.1 $\rightarrow$ 74.7\% \\
\bottomrule
\end{tabularx}
\end{table*}

\subsubsection{Training and evaluation}
Training uses Isaac Lab \cite{mittal2025isaaclab} with two policy-controlled carriers and one rigid payload. Policy, guidance, and WBC rates are 2, 0.5, and 100~Hz. Each scenario uses common success, termination, and horizon rules. SR is evaluated over 100 episodes per seed, and family SR averages its three scenarios. Grasp loss or payload drop is terminal. We report mean $\pm$ standard deviation over five seeds.

Training uses Adam with cosine annealing and orthogonal initialization for $2.0\times10^9$ action steps. Table~\ref{tab:training_hyper} lists the shared settings. Optimizer comparisons fix task definitions, the 210-D input, 11-D residual action, reward, geometry checks, WBC, evaluation budget, task gains, and QP weights.

Robot-script uses independent proximal policy optimization (IPPO) against a scripted kinematic collaborator. HAPPO, HATRPO, and PPO+PCGrad use two adaptive policy-controlled carriers. Robot-script uses the same [256,256,128] multilayer perceptron (MLP) backbone and $2.0\times10^9$-step budget. Its partner proxy supplies relative position, velocity, and orientation with 2.0~cm Gaussian vertical jitter, 0.1~m/s uniform velocity noise, and 3.0$^\circ$ Gaussian yaw noise.

All configurations retain the task-space action and execution contracts. Table~\ref{tab:giant_matrix} therefore compares complete coordination configurations under the same C2C interfaces. For the simulation-to-human study, MARL training uses two adaptive policy-controlled carriers. At hardware deployment the learned robot actor is frozen. The second learned carrier is removed and the human's motion-capture (MoCap) state fills the same partner fields. The human is not optimized by MARL and no online learning is performed during the physical trials. For scenario $s$, $\overline{\mathrm{SR}}_{\mathrm{learn},s}$ averages HAPPO, HATRPO, and PPO+PCGrad. Table~\ref{tab:giant_matrix} reports $\Delta_s=100(\overline{\mathrm{SR}}_{\mathrm{learn},s}-\mathrm{SR}_{\mathrm{script},s})/\mathrm{SR}_{\mathrm{script},s}$. The mean row applies this definition to the method means.

\subsection{Coordination algorithms under fixed contracts}
We compare HAPPO \cite{zhong2022heterogeneous}, HATRPO \cite{kuba2021trust}, and PPO+PCGrad with robot-script IPPO. The adaptive MARL variants use different updates but share fixed task and execution interfaces. Robot-script is a fixed-partner reference. This tests whether coordination algorithms can change without redesigning the surrounding HRC system.

Mean SR is 56.5\%, 77.1\%, 80.8\%, and 82.1\% for robot-script, HAPPO, HATRPO, and PPO+PCGrad. Every adaptive variant improves over robot-script in all nine scenarios while preserving the same payload reference, observation fields, 11-D residual command, reward, geometry checks, and WBC. The largest relative gains occur for long-object handling, where rigid-payload coupling amplifies timing and heading mismatch. No adaptive optimizer leads every scenario. This is precisely the intended separation: coordination performance changes while the HRC task and execution interface remain fixed. Table~\ref{tab:giant_matrix} and Fig.~\ref{fig:return_and_tables}(a)--(b) report the results.

\subsection{Simulation-to-human transfer under the same contract}
\begin{figure*}[t]
    \centering
    \includegraphics[width=0.99\textwidth]{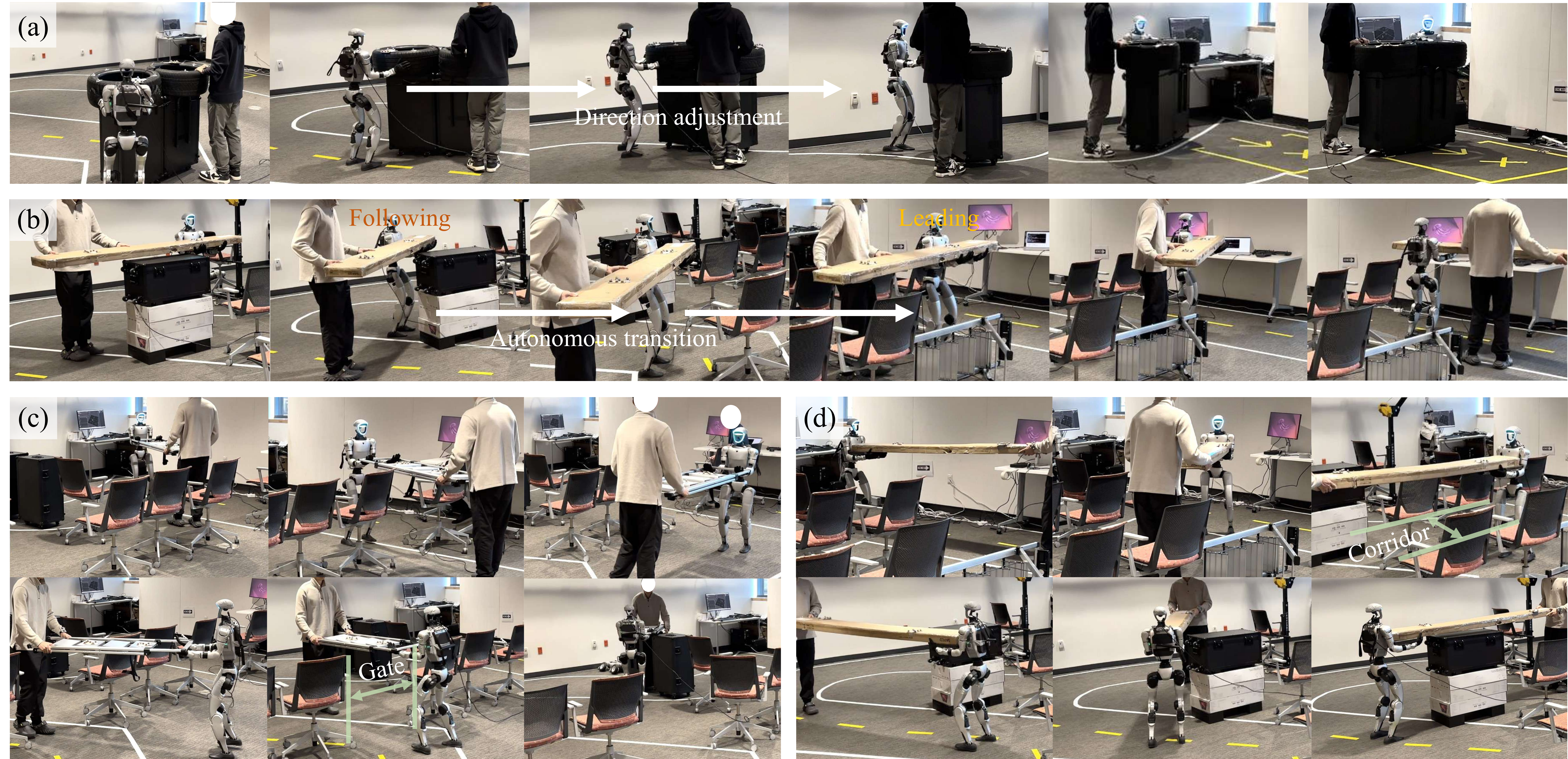}
    \caption{Physical human--single robot collaboration demonstrations (a) online heading adjustment, (b) following-to-leading transition, (c) constrained gate traversal, and (d) super-long-payload corridor transport.}
    \label{fig:real_world_overview}
\end{figure*}

\begin{figure}[t]
    \centering
    \includegraphics[width=0.99\columnwidth]{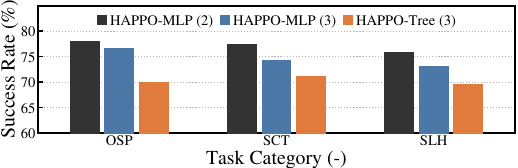}
    \caption{Task-family success rate for two-carrier HAPPO-MLP, three-carrier HAPPO-MLP, and three-carrier HAPPO-Tree. Parentheses give carrier count.}
    \label{fig:three_compare}
\end{figure}

\begin{figure}[t]
    \centering
    \includegraphics[width=0.99\columnwidth]{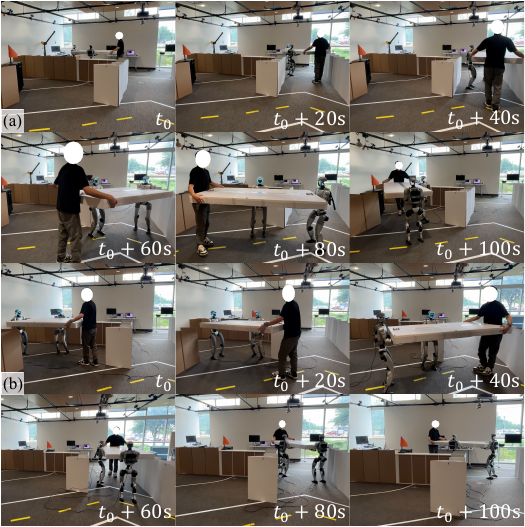}
    \caption{Physical human--multiple robots collaborative transport with two Unitree G1 humanoids and one human: (a) payload reorientation and (b) passage traversal around laboratory partitions, shown at 20~s intervals.}
    \label{fig:three_task}
\end{figure}
Physical tests use five trials per task with common guidance and robot-side WBC. Measured human motion fills the same partner-state fields used by simulated carriers, and the frozen robot actor retains the same residual base/wrist action space. The partner source changes without changing the robot policy interface.

Compared with robot-script IPPO, PPO+PCGrad retains 100\% SR on OSP, raises SCT from 40\% to 100\%, and raises SLH from 40\% to 80\%. Completion time decreases from 101.6 to 81.5~s on SCT and from 113.6 to 86.9~s on SLH, with lower tilt rate in both tasks. Table~\ref{tab:real_stats_updated} and Fig.~\ref{fig:return_and_tables}(c) report the results. Fig.~\ref{fig:real_world_overview} shows heading adjustment, following-to-leading transition, gate traversal, and long-payload transport.

\begin{table}[h]
\centering
\caption{Physical results over five trials per task.}
\label{tab:real_stats_updated}
\scriptsize
\setlength{\tabcolsep}{2.0pt}
\renewcommand{\arraystretch}{1.15}
\begin{tabular*}{\columnwidth}{@{\extracolsep{\fill}} l ccc ccc}
\toprule
\multirow{2}{*}{\textrm{Task}} & \multicolumn{3}{c}{\textrm{Robot-script IPPO}} & \multicolumn{3}{c}{\textrm{PPO+PCGrad}} \\
\cmidrule{2-4} \cmidrule{5-7}
& SR & Tilt ($^\circ$/s) & Time (s) & SR & Tilt ($^\circ$/s) & Time (s) \\
\midrule
OSP & 100\% & -- & 74.2 & 100\% & -- & 65.2 \\
SCT & 40\% & 3.2 & 101.6 & 100\% & 2.4 & 81.5 \\
SLH & 40\% & 3.6 & 113.6 & 80\% & 2.7 & 86.9 \\
\bottomrule
\end{tabular*}
\end{table}

\subsection{Policy representation and carrier-count extension}
We next test whether the coordination contract is tied to one function approximator or one fixed team size. Under HAPPO, multilayer perceptron actors are tested with two and three carriers, and a hard oblique decision-tree actor is tested with three. For an internal node $k$, the tree uses the hard split $g_k(o)=\mathbb{I}[w_k^T o>b_k]$. The selected leaf $l(o)$ stores an affine controller and outputs
\begin{equation}
a(o)=a_c+a_s\odot\tanh\!\left(K_{l(o)}o+b_{l(o)}\right),
\label{eq:tree_policy}
\end{equation}
where $a_c$ and $a_s$ are the center and half-range of the action bounds, and $K_l,b_l$ are leaf parameters. Eq.~\eqref{eq:tree_policy} gives a bounded action with an explicit root-to-leaf path. Forward routing is hard and deterministic, while a soft backward gate optimizes split parameters during training. The tree uses the same physical inputs and bounded 11-D residual action as the neural actor, making its deployed decision path directly inspectable.

For three carriers, only partner encoding changes inside coordination. Task guidance, action semantics, reward, and WBC remain fixed, and the MLP/tree comparison uses the same three-carrier adapter. This isolates representation and carrier-count changes from cognition and execution.

Mean SR is 77.1\% for two-carrier HAPPO-MLP, 74.7\% for three-carrier HAPPO-MLP, and 70.3\% for three-carrier HAPPO-Tree. Moving from two to three MLP carriers changes the mean by only 2.4 points, while replacing the three-carrier MLP with the interpretable tree costs 4.4 points without changing the task or action contract. Fig.~\ref{fig:three_compare} summarizes this controlled comparison. Fig.~\ref{fig:three_task} shows two Unitree G1 humanoids and one human performing payload reorientation and passage traversal, closing the three-carrier interface on hardware.

Together, the experiments validate contract preservation across coordination algorithm, policy representation, carrier count, and partner-state source. The deterministic fallback also exercises the cognition boundary by producing the same $P_t$ when VLM proposals are rejected. Quantitative simulations retain the reference VLM and WBC, while hardware closes the sim-to-human and three-carrier interfaces. Across these tested substitutions, neighboring modules retain the same physical semantics and robot-side feasibility checks.

\section{Conclusion}

This work identifies interface semantics as a central systems challenge in full-stack physical HRC. Rather than coupling cognition, partner adaptation, learned coordination, and whole-body control through implementation-specific representations, C2C organizes them around physical contracts defined on the shared payload. The resulting architecture preserves the collaboration task while allowing individual modules to change behind stable interfaces, providing a principled basis for composable human--humanoid collaboration.

\begin{itemize}
\item The proposed object-centric contracts separate task intent, partner state, learned coordination, and robot execution through physically interpretable interfaces, preventing changes within one module from propagating across the full stack.

\item This contract structure supports heterogeneous implementations behind the same interfaces, including different MARL optimizers, neural and hard-tree policies, simulated and human partners, and two- and three-carrier configurations, while retaining common task and action semantics.

\item Across nine simulated scenarios, adaptive MARL achieves 77.1--82.1\% mean success compared with 56.5\% for the robot-script reference. Physical G1--human experiments achieve 100\%, 100\%, and 80\% success on OSP, SCT, and SLH, while the three-carrier system retains 74.7\% success with HAPPO-MLP and 70.3\% with the hard-tree policy, with further validation on two G1 humanoids cooperating with one human.
\end{itemize}

\bibliographystyle{IEEEtran}
\bibliography{IEEEexample_reviewer_hardened}

\end{document}